\documentclass[letterpaper, 10 pt, conference]{ieeeconf}  

\IEEEoverridecommandlockouts                              

\usepackage{graphics} 
\usepackage{epsfig} 
\usepackage{mathptmx} 
\usepackage{amsmath} 
\usepackage{amssymb}  
\usepackage{array}
\usepackage{booktabs} 
\usepackage{tabularx}
\usepackage{makecell}
\usepackage{multirow}

\title{\LARGE \bf
Khalasi: Energy-Efficient Navigation for Surface Vehicles \\in Vortical Flow Fields
}

\author{Rushiraj Gadhvi$^{1}$ and Sandeep Manjanna$^{1}$
\thanks{$^{1}$ Rushiraj Gadhvi and Sandeep Manjanna both are with Plaksha University, Mohali, India. {\tt\small gadhvirushiraj@gmail.com}; \tt\small sandeep.manjanna@plaksha.edu.in}}%

\usepackage{hyperref}
\usepackage{float}
\begin{document}

\maketitle
\thispagestyle{empty}
\pagestyle{empty}

\begin{abstract}
For centuries, \emph{khalasi} (Gujarati for sailor) have skillfully harnessed ocean currents to navigate vast waters with minimal effort. Emulating this intuition in autonomous systems remains a significant challenge, particularly for Autonomous Surface Vehicles (ASVs) tasked with long-duration missions under strict energy budgets. In this work, we present a learning-based approach for energy-efficient surface vehicle navigation in vortical flow fields, where partial observability often undermines traditional path-planning methods. We present an end-to-end reinforcement learning framework based on Soft Actor–Critic (SAC) that learns flow-aware navigation policies using only local velocity measurements. 
Through extensive evaluation across diverse and dynamically rich scenarios, our method demonstrates substantial energy savings and robust generalization to previously unseen flow conditions, offering a promising path toward long-term autonomy in ocean environments. The navigation paths generated by our proposed approach show an improvement in energy conservation $30\% - 50\%$ compared to the existing state-of-the-art techniques.
\end{abstract}

\section{INTRODUCTION}

Ocean surveillance plays a critical role in a wide range of applications, including environmental monitoring~\cite{Mukhopadhyay2014}, oceanographic surveying~\cite{Smith2010}, biodiversity tracking~\cite{tittensor2010global, KUHNZ2020104761}, and climate change research~\cite{smith2021abyssal, WORM2021445}. Wave gliders, underwater gliders, and active drifters are often used for such tasks for longer periods while operating on limited energy budgets~\cite{jmse13040717, Grare2021}. 
In these missions, autonomous vehicles must navigate through unstable, turbulent wind, or ocean currents. A smart robot can actually take advantage of these surrounding flows, using them to move toward its goal while conserving energy.

Traditional path planning methods are often unsuitable for these dynamic environments. For instance, optimal control solutions to Zermelo's navigation problem~\cite{zermelo-rl} require full knowledge of the entire flow-field, which is impractical to obtain in the real world. Similarly, the drawback with the use of traditional graph-based methods such as Dijkstra, A*, particle swarm optimization (PSO) is that it considers the flow-field to be time invariant, which is not the case in ocean currents. Moreover, the robot may not have access to the entire flow-field and must therefore act autonomously, relying solely on data from its onboard sensors.

In recent years, deep-reinforcement learning (DRL) has emerged as a promising alternative to traditional control and planning methods for the navigation of Autonomous Surface and Underwater Vehicles (ASVs/AUVs), particularly in environments with limited prior knowledge. The main advantage of RL (Reinforcement Learning) is its ability to learn navigation policies in simulation, effectively addressing the data scarcity challenge posed by complex ocean dynamics. The availability and quality of the real-world flow field data is often limited, making simulation-based training a scalable and practical approach to policy development.

Through simulations, RL-based approaches have been successfully used to create energy-efficient swimming gaits for robotic fish~\cite{energy-efficient-fish, energy-efficient-fish2}, to navigate vortical fields~\cite{og-dabiri, sensing-flow}, for the navigation of microbots and nanobots in medical applications~\cite{micro-nano-robots}, and to replicate gaits of microorganisms~\cite{micro-orgamism-gait}. A key benefit of RL across application domains is its capacity to develop robust policies that generalize well to new, unseen conditions.

In this work, we propose \emph{khalasi} (Gujarati for sailor) an end-to-end pipeline for ASVs to navigate energy efficiently in cortical flow fields. Our contributions in this work include the following.

\begin{itemize}
\item Policy architecture that relies solely on local velocity observations, enabling robust operation in partially observable and sensing-constrained environments.
\item A thoroughly evaluated generalized learning pipeline that exhibits significant energy conservation in ASVs navigating in unseen complex flow fields, including real ocean currents.
\item A benchmark simulation set-up for evaluating planning algorithms in vortical flow environments.
\end{itemize}

To facilitate future research, we have open-sourced our project at: \url{https://gadhvirushiraj.github.io/khalasi}.


\begin{figure*}[t]
  \centering
  \includegraphics[width=0.8\textwidth]{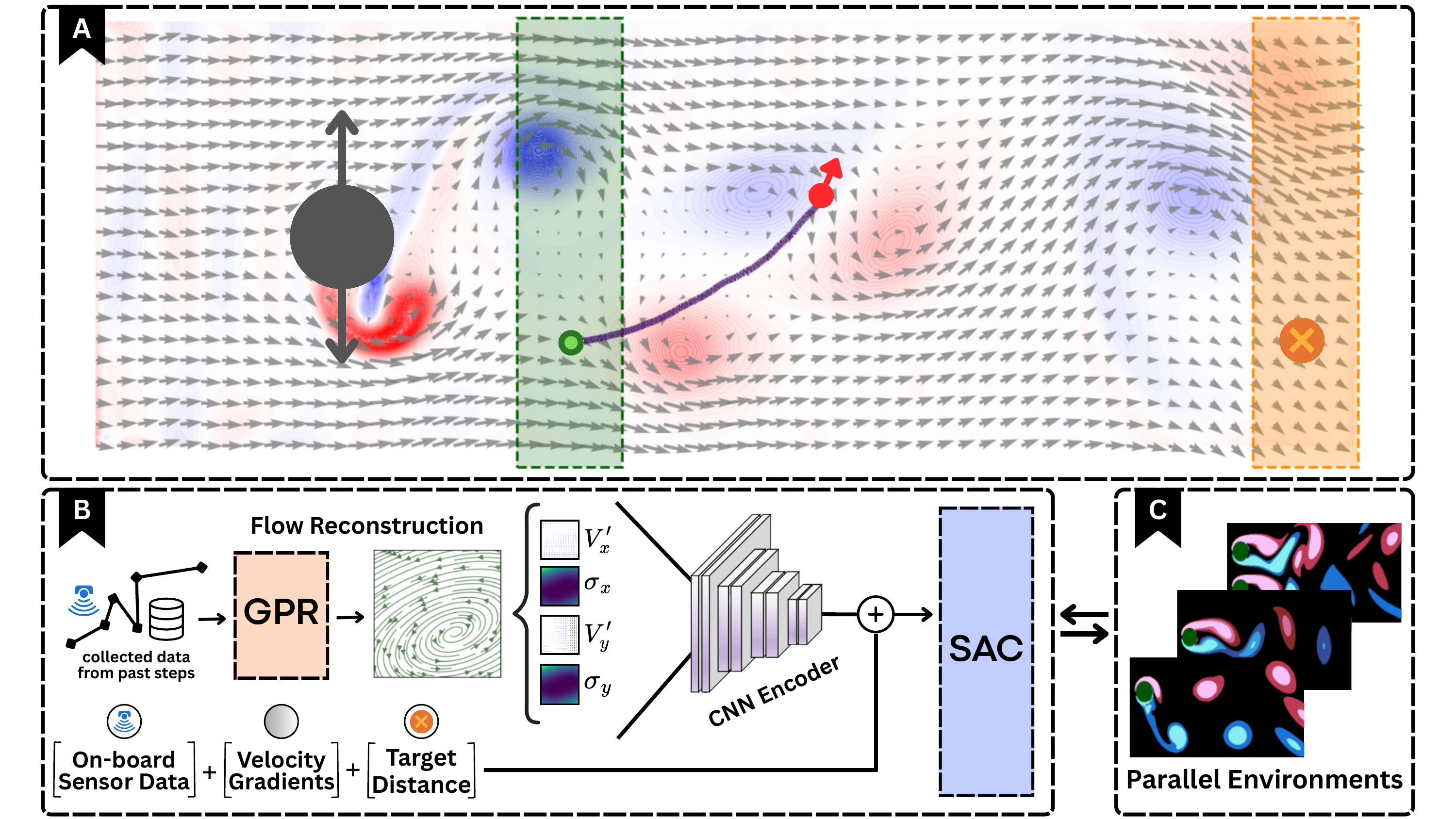}
  \caption{Overview of \textit{khalasi} pipeline. \textbf{(A)} Training and evaluation environment: The agent (red) navigates toward the goal (orange) by leveraging local flow features. The green box and orange box denotes the agent and goals spawn region respectively. \textbf{(B)} Training pipeline: historical local velocity observations are processed through a Gaussian Process Regression (GPR) module for flow reconstruction, followed by spatial gradient extraction via a CNN encoder. The resulting latent representation, combined with positional and goal data, is input to a Soft Actor–Critic (SAC) policy. \textbf{(C)} Trained using parallelized environments with diverse flow variations enable better generalization across different flows.}
  \label{fig:overview}
  \vspace{-1.5em}
\end{figure*}

\section{RELATED WORKS}

Recent work has applied RL to navigation in dynamic flow fields as an alternative to model-based control. Biferale et al.~\cite{zermelo-rl} extended RL to Zermelo’s navigation problem, showing robustness in turbulent regimes but relying on long-range flow data, which limits applicability in scenarios with restricted sensing. Gunnarson et al.~\cite{og-dabiri} demonstrated that agents using local velocity signals outperformed bioinspired vorticity models, although the policies exhibited limited generalization. Jiao et al.~\cite{sensing-flow} showed that egocentric sensing improved generalization and explored velocity gradient sensing inspired by seal whiskers. Zhang et al.~\cite{flow-prediction} added short-term flow prediction via Long Short-Term Memory (LSTM), but performance degraded on longer horizons and required large training data to handle the variability inherent in real-world ocean dynamics.

Classical methods for energy-efficient navigation use search-based and model-driven strategies. Garau et al.~\cite{trad-energy-eff-garau} applied A* with ocean currents, Koay et al.~\cite{trad-energy-eff-kaoy} added obstacle avoidance, and Lee et al.~\cite{trad-energy-eff-lee} integrated tidal currents and bathymetry. Niu et al.~\cite{trad-energy-eff-niu} combined Voronoi diagrams with visibility graphs for current-aware trajectories, but most of these assume global flow knowledge and temporal stability, which rarely hold in practice~\cite{trad-energy-eff-niu2}. Other refinements include smoother Riemannian-based paths~\cite{anis-work}, adaptive A* replanning~\cite{zhangs-work}, and bio-inspired or RL methods such as vortex-riding fish strategies~\cite{fish-verma} and cyclic PPO trajectories~\cite{mei2023}. While effective, these still face limits in dynamic environments, motivating adaptive, data-driven approaches like reinforcement learning.

Compared to prior work, Khalasi introduces a more robust approach to energy-efficient navigation by relying only on single-point sensor measurements rather than full flow-field knowledge. By training across diverse environments, the model generalizes well to unseen flow conditions, as shown through extensive testing. Leveraging reinforcement learning, the agent develops an intuition for detecting and interpreting flow patterns, enabling it to harness currents effectively and navigate efficiently toward its target.
 



\section{Problem Formulation}\label{Sec:problem_formulation}

We consider the problem of navigating an agent toward a target location in a time-varying, vortical flow field with the objective of discovering a quasi-optimal energy-efficient trajectory. To simulate background turbulence, we generate 2D unsteady flow fields using variations of von Kármán vortex street configurations, as illustrated in Fig.~\ref{fig:overview}(C). These include single-cylinder, double-cylinder, and oscillating-cylinder setups, which are described in detail in Section~\ref{sec:algo-train}. The local flow velocity at any point $(x,y)$ is denoted by $\mathbf{v}_{\text{flow}}(x, y, t) \in \mathbb{R}^2$ and varies both spatially and temporally. The agent and target are randomly initialized within designated regions of the environment, as shown in Fig.~\ref{fig:overview}(A), where the green region indicates the agent’s spawn area and the orange region the target’s spawn area. A navigation episode is considered successful if the agent reaches the target without exiting the environment or exceeding a fixed time limit; otherwise it is marked as a failure.

The target is positioned at $\mathbf{p^*} \in \mathbb{R}^2$. The agent is modeled as a point mass located at position $\mathbf{p} \in \mathbb{R}^2$ and with velocity $\mathbf{u} \in \mathbb{R}^2$, solely induced by propulsion. The agent controls two imbalanced thrusters ($T_{l}$ and $T_{r}$) with ranges of $T$ N (forward) and $T/2$ N (reverse) to move within the environment, but it lacks direct control over its direction of travel. The value of $T$ is chosen empirically based on the maximum flow velocity observed across the environments. For simplicity, the agent's movement does not affect the flow dynamics. The onboard sensors are assumed to be accurate and provide real-time access to local flow velocity and absolute position. No global map or flow forecast is available, decisions are made purely based on local sensing. We build a relation between the thrust controls and the energy consumption for our evaluations as described below.
\vspace{-0.4em}
\begin{align*}
E &= \int \left(V_{l}^2 + V_{r}^2 \right) dt \\
  &\approx {V_{in}^2} \Delta t \sum \left(L_{cmd}^2 + R_{cmd}^2\right)\\[4pt]
  &\propto \sum \left(T_{l} + T_{r}\right)
\end{align*}
\vspace{-0.3em}
where $V_{l}$ and $V_{r}$ are the voltages applied to the left and right motors, respectively, $V_{in}$ denotes the battery voltage, and $\Delta t$ represents the duration between successive action commands. $L_{cmd}$ and $R_{cmd}$ represent the PWM signals sent to the left and right motors, respectively. The terms $T_{l}$ and $T_{r}$ refer to the thrust generated by the respective motors, which are directly determined by the action output of our agent.

\section{METHODOLOGY}

\subsection{Flow Reconstruction}
Relying solely on instantaneous single-point data is akin to driving a car while peeking through a keyhole. An agent's decision-making capacity is closely tied to the quality and extent of available observations from the environment. In our setup, the key limitation is not the quality of the data, but the amount, since the agent can only sense at a single point. To overcome this, we exploit the temporal aspect of the sensor data to get a better sense of the surrounding flow structure.

\begin{figure}[h]
  \centering
  \includegraphics[width=0.8\columnwidth]{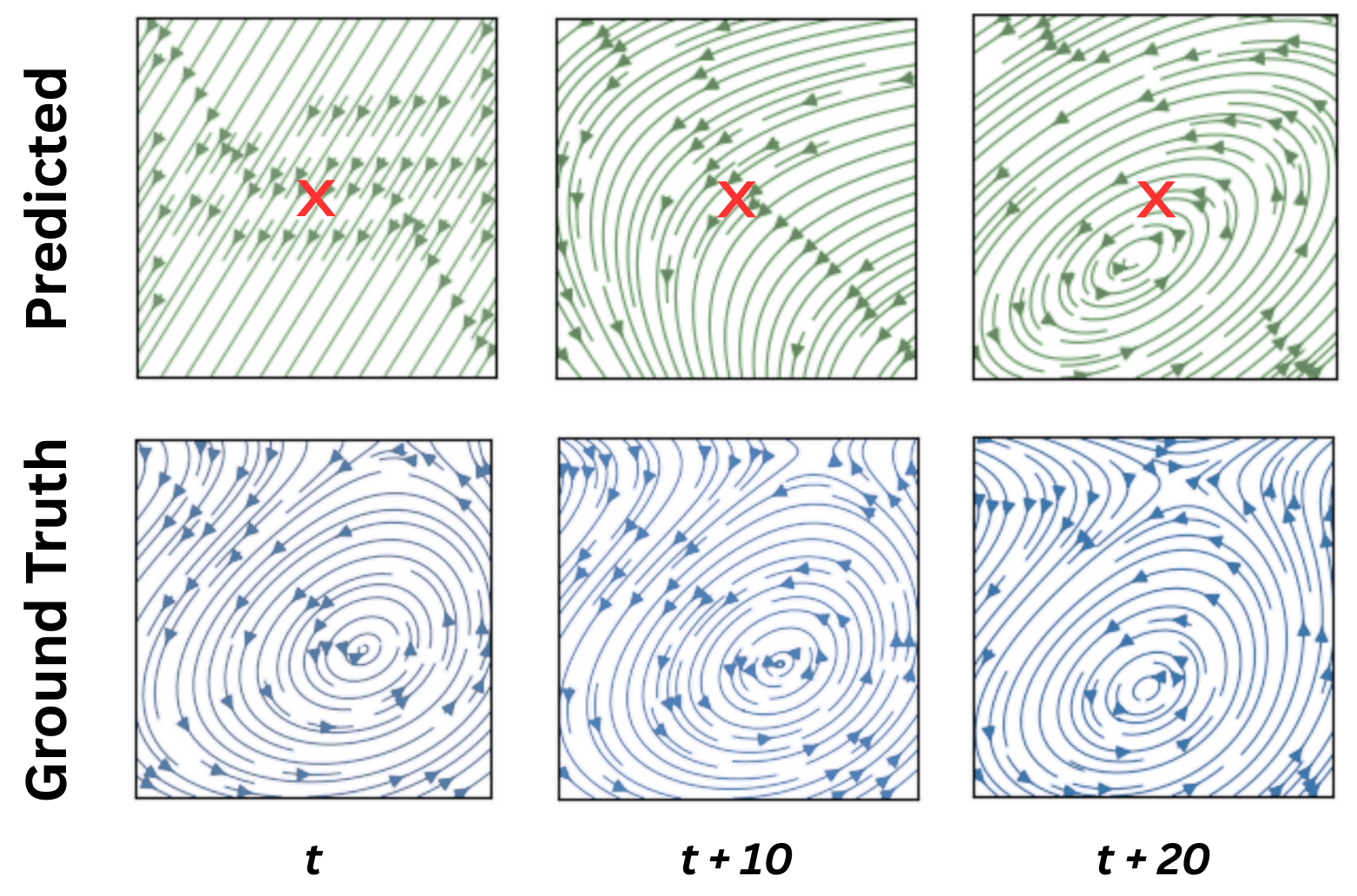}
  \vspace{-0.5em}
  \caption{Example of flow reconstruction in a vortex environment using GPR. Accuracy improves as more temporal samples are incorporated, shown by predictions at $t$, $t+10$, and $t+20$ compared to the ground truth.}
  \label{fig:gpr}
  \vspace{-0.5em}
\end{figure}

We employ Gaussian Process Regression (GPR)~\cite{gpr} to reconstruct the local velocity field from a sequence of observations. GPR is a nonparametric Bayesian method that represents an unknown function as a distribution over possible functions. In contrast to neural network approaches such as LSTM~\cite{lstm} or Physics-Informed Neural Networks (PINNs)~\cite{pinn}, which require large training data and may struggle to generalize, GPR offers a better balance between accuracy and generalization. It is also more suited for real-world deployment, as it does not require significant compute power, unlike other methods. In our setup, the GPR operates on a sliding window of the past $t_{limit}$ sensor measurements. From these temporal samples, a continuous flow field is estimated for a $64 \times 64$ agent-centric grid. We use a Radial Basis Function (RBF) kernel for GPR, to capture the smooth and continuous nature of ocean currents. This reconstructed local flow map provides a richer representation of the surrounding dynamics, allowing the agent to make informed decisions.

\begin{figure}[h]
  \centering
  \includegraphics[width=0.75\columnwidth]{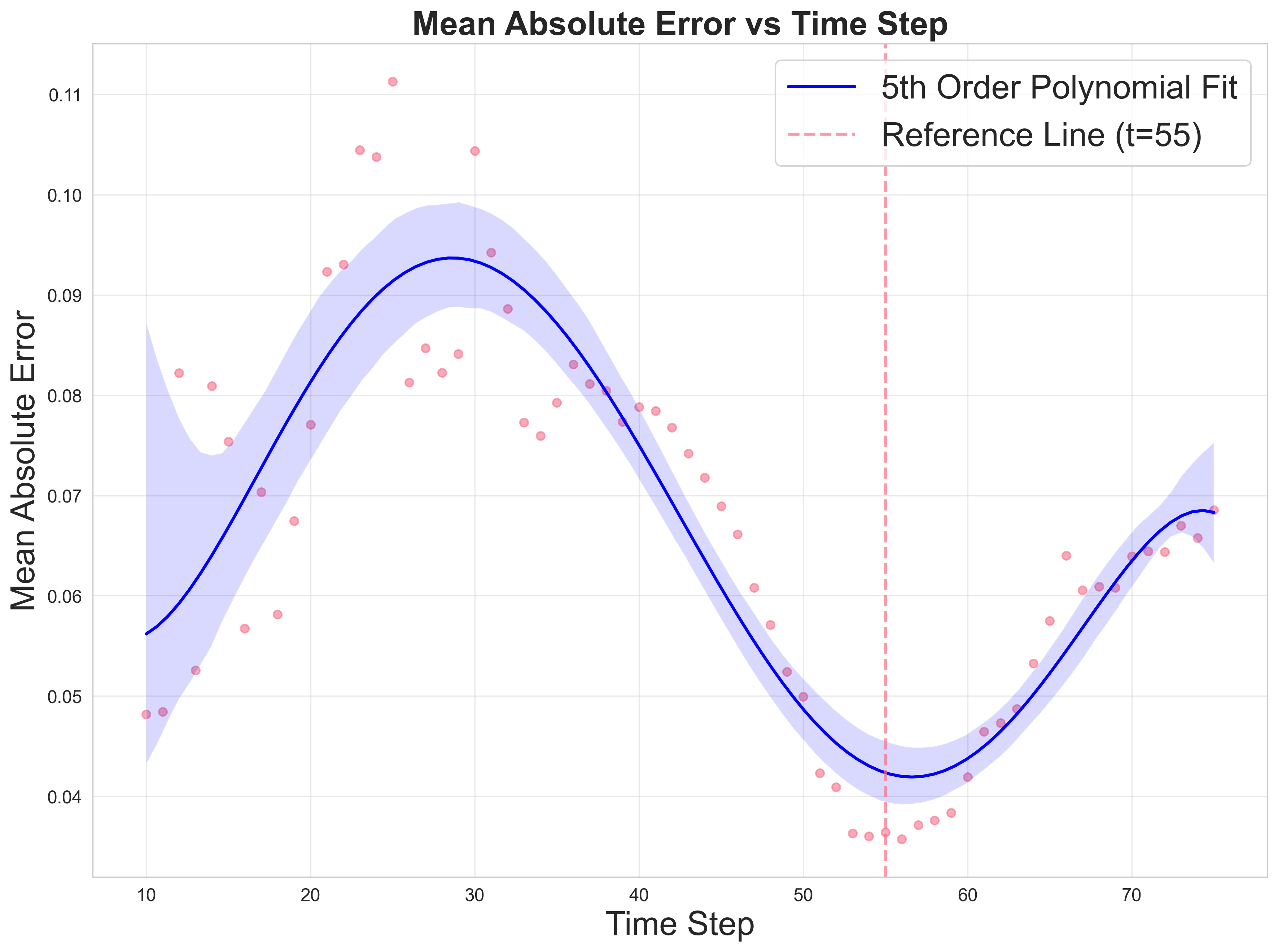}
  \caption{Mean Absolute Error (MAE) of GPR-based flow reconstruction compared to ground truth, evaluated along a dummy agent trajectory in the same environment.}
  \label{fig:mae_gpr_plot}
  \vspace{-0.5em}
\end{figure}

The hyperparameter $t_{limit}$ controls the number of past measurements used for GPR. Its value depends on how dynamic the flow is. The value for $t_{limit}$ was chosen through empirical experiments illustrated in Fig.~\ref{fig:mae_gpr_plot}. Based on the mean absolute error, we use $t_{limit} = 55$ in our experiments. The GPR estimates degrade after this limit as older measurements are no longer indicative of the temporal field.

\subsection{Observation and Action Space}
Observation is divided into two parts ($O_1$ and $O_2$). First, GPR estimates the fields $V'_x$ and $V'_y$ centered around the agent’s position $\mathbf{p}_t$ at time $t$. It also computes the cell-wise standard deviations $\sigma_x$ and $\sigma_y$ to capture uncertainty in the estimates. We stack these four maps into a tensor $\mathbf{X}_t \in \mathbb{R}^{64 \times 64 \times 4}$:
\[
\mathbf{X}_t = \mathrm{stack}\big(V'_x(\mathbf{p}_t, t),\, V'_y(\mathbf{p}_t, t),\, \sigma_x(t),\, \sigma_y(t)\big)
\]
\noindent
A convolution neural network (CNN) encoder maps this tensor to a latent vector $\mathbf{z}_t \in \mathbb{R}^{d}$. This forms our $O_1$.
\[
O_1 = \mathbf{z}_t = \phi_{\text{CNN}}(\mathbf{X}_t)
\]

\noindent
The second part of the observation is a low-dimensional state vector:
\begin{equation*}
\begin{aligned}
O_2 = [&\, \Delta x_t,\, \Delta y_t,\, u_x(t),\, u_y(t),\, v_x(\mathbf{p}_t, t),\, v_y(\mathbf{p}_t, t), \\
      &\, \nabla \cdot u_x(t),\, \nabla \cdot u_y(t),\, \theta_t \,]
\end{aligned}
\end{equation*}

Where, $\Delta x_t$ and $\Delta y_t$ represent the relative position to the goal, $u_x(t)$ and $u_y(t)$ are the velocity components of the agent, and $v_x(\mathbf{p}_t, t)$ and $v_y(\mathbf{p}_t, t)$ are the local flow velocity components at the agent’s position. The terms $\nabla \cdot \mathbf{v}(\mathbf{p}_t, t)$ and $\nabla \cdot \mathbf{v}(\mathbf{p}_t, t)$ represent the local flow gradients. Finally, $\theta_t$ denotes the agent’s heading, which—unlike prior work~\cite{og-dabiri, flow-prediction}—is not directly controlled. Finally, the complete observation provided to the policy is the concatenation of the two parts:$o_t = [O_1,\, O_2]$.

We consider a continuous action space where the action at time $t$ is denoted as $\mathbf{a}_t = [a^l_t,\, a^r_t] \in \mathbb{R}^2$ and represents the thrust values applied by each thruster. The action values are in the range $[-1, +1]$. As mentioned in~\ref{Sec:problem_formulation}, the thrusters are unbalanced. The action values are mapped to the range $[-T/2, T]$ before being executed on the agent.

\subsection{Rewards}

The reward function is designed to guide the agent toward its goal while balancing energy efficiency and smooth control. At each time step $t$, the total reward is a weighted sum of five terms:
\[
R_t = w_1 r_{\text{target}} + w_2 r_{\text{dist}} + w_3 r_{\text{thrust}} + w_4 r_{\text{surf}} + w_5 r_{\text{jitter}}
\]
where $w_i$ are positive weights that control the influence of each component. The values for these weights are application-specific and for our current work, we use a unit weight. Each of the reward terms is discussed below, with $C_{1-4}$ representing constants.


\subsubsection{Target Reward}
A sparse bonus $r_{\text{target}}$ is given when the agent reaches the target. This occurs if the agent’s position ($\mathbf{p}_t$) is within a radius ($\epsilon$) of the target ($\mathbf{p}^*$):
\[
r_{\text{target}} =
\begin{cases}
    C_1 & \text{if } ||\mathbf{p}_t - \mathbf{p}^*||_2 < \epsilon, \\
    0 & \text{otherwise}.
\end{cases}
\]


\subsubsection{Distance Reward}
To provide a denser learning signal, $r_{\text{dist}}$ rewards the agent for reducing the distance to the target compared to the previous time step. Displacement-based rewards are not used because they indirectly prioritize time efficiency, which is not the main focus of this work. We found this modification to be crucial for enabling the agent to exhibit wave-riding behavior.
\[
r_{\text{dist}} =
\begin{cases}
    C_2 & \text{if } ||\mathbf{p}_{t-1} - \mathbf{p}^*||_2 > ||\mathbf{p}_t - \mathbf{p}^*||_2, \\
    -C_2 & \text{otherwise}.
\end{cases}
\]


\subsubsection{Thruster Penalty}
To promote energy efficiency, $r_{\text{thrust}}$ penalizes by a magnitude proportional to thruster use. The penalty is proportional to the L1 norm of the action vector $\mathbf{a}_t = [a^l_t, a^r_t]^T$:
\[
r_{\text{thrust}} = -C_3||\mathbf{a}_t||_1 = -C_3(|a^l_t| + |a^r_t|).
\]


\subsubsection{Surfing Reward}

The surfing reward $r_{\text{surf}}$ encourages the agent to take advantage of the surrounding flow and maintain velocity while using as little thrust as possible. The reward is computed as a linear function of thrust usage, provided that the agent’s velocity exceeds the local flow velocity and the applied thrust remains within defined limits. 
Formally, the reward is given by:
\[
r_{\text{surf}} =
\begin{cases}
    R_{\text{max}} - ||\mathbf{a}_t||_1 \cdot
    \dfrac{R_{\text{max}} - R_{\text{min}}}{U_{\text{max}}} 
    & \text{if } 
    \begin{aligned}[t]
        & ||\mathbf{a}_t||_1 \le U_{\text{max}} \\
        & \text{and } \mathbf{u_t} > \textbf{v}(\mathbf{p}_t, t)
    \end{aligned} \\[8pt]
    0 & \text{otherwise}.
\end{cases}
\]

Here, $U_{\text{max}}$ represent the maximum thrust levels, while $R_{\text{min}}$ and $R_{\text{max}}$ represent the minimum and maximum reward values. These parameters are set empirically based on the characteristics of the system.


\subsubsection{Jitter Penalty}
To encourage stable control, $r_{\text{jitter}}$ penalizes large changes in thruster commands in consecutive time steps:
\[
r_{\text{jitter}} = -C_4 \cdot ||\mathbf{a}_t - \mathbf{a}_{t-1}||_1,
\]
where $\mathbf{a}_{t-1}$ is the action chosen in the previous time step.


\begin{figure}[h]
  \centering
  \includegraphics[width=\columnwidth]{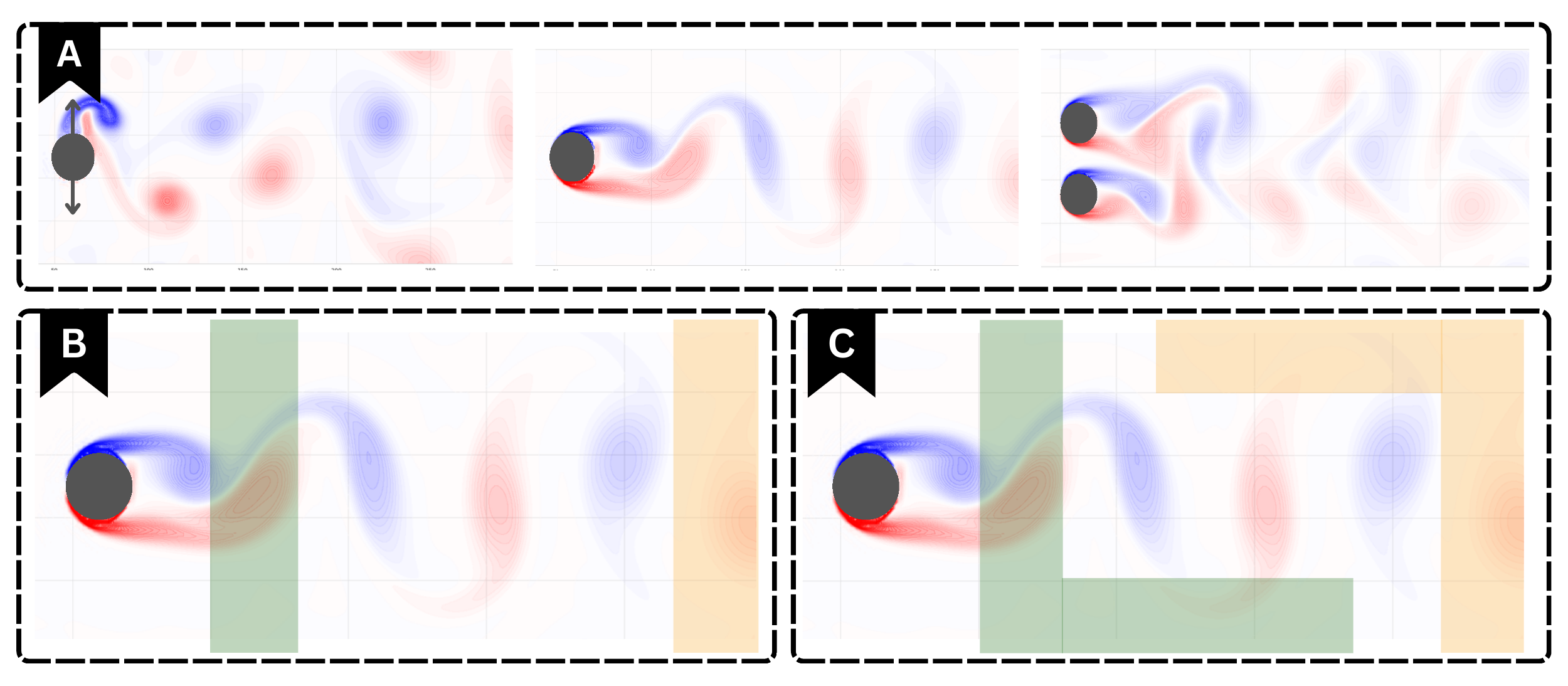}
  \vspace{-0.5em}
    \caption{\textbf{(A)} Environment setups shown from left to right: oscillating single cylinder, static single cylinder, and static double cylinder. \textbf{(B,C)} Spawn regions used for training and testing: vertical spawn and L-shaped spawn, respectively.}
  \label{fig:setup}
  \vspace{-1em}
\end{figure}

\subsection{RL Algorithm and Training} \label{sec:algo-train}

We use Soft Actor–Critic (SAC)~\cite{SAC} as the reinforcement learning algorithm to train the energy efficient navigation policy. SAC is well known for its unique entropy-based exploration, which is crucial for discovering complex trajectories that utilize flow for energy efficiency. Through our trials, we found that SAC outperformed both the Proximal Policy Optimization (PPO)~\cite{PPO} and the Twin Delayed Deep Deterministic Policy Gradient (TD3)~\cite{TD3} in our setup.

The SAC model is trained on three flow environments (illustrated in Fig.~\ref{fig:setup}(A)) in parallel. The first is an oscillating cylinder located at $(W/5, H/2)$, moving vertically with an amplitude of $H/6$ and a period of 500 steps. The second is a single cylinder at $(W/5, H/2)$ that generates a simple von Kármán vortex street. The third is a double-cylinder set-up with cylinders positioned at $(W/5, H/3)$ and $(W/5, 2H/3)$. Here, $W$ and $H$ represent the width and height of the environment, which are set to $300 \times 100$ units. Training runs for each environment are run independently and experiences are collected asynchronously. These experiences are used to perform batch updates on the policy. Training across multiple environments exposes the agent to diverse flow patterns, enhancing its generalization and improving the training stability.

In our experiments, we introduce two layouts for spawning the start and the target location for the agent. The spawn layout presented in Fig.~\ref{fig:setup}(B) is referred to as vertical random spawn and that in Fig.~\ref{fig:setup}(C) is referred to as L-shaped random spawn. The vertical spawn layout was used during training and testing, while the L-shaped layout was used only for testing. This is because it generates scenarios where the boat must apply high thrust and navigate against the current. Although our problem does not focus on the time taken to reach the goal, a hard limit on the number of steps is necessary for training. To determine this limit, we measured the number of steps an agent takes to drift from the left to the right side of the environment without any thrust. On average, this took 1200 steps. We added $25\%$ additional steps to account for variability and set the episode limit to 1500 steps. If an agent exceeds this limit, we conclude that it failed to reach the goal.

The model was trained for $2 \times 10^6$ steps. Training was performed on an NVIDIA RTX A5000 and took approximately 16–18 hours including test runs.


\begin{figure*}[!ht]
  \centering
  \includegraphics[width=0.75\textwidth]{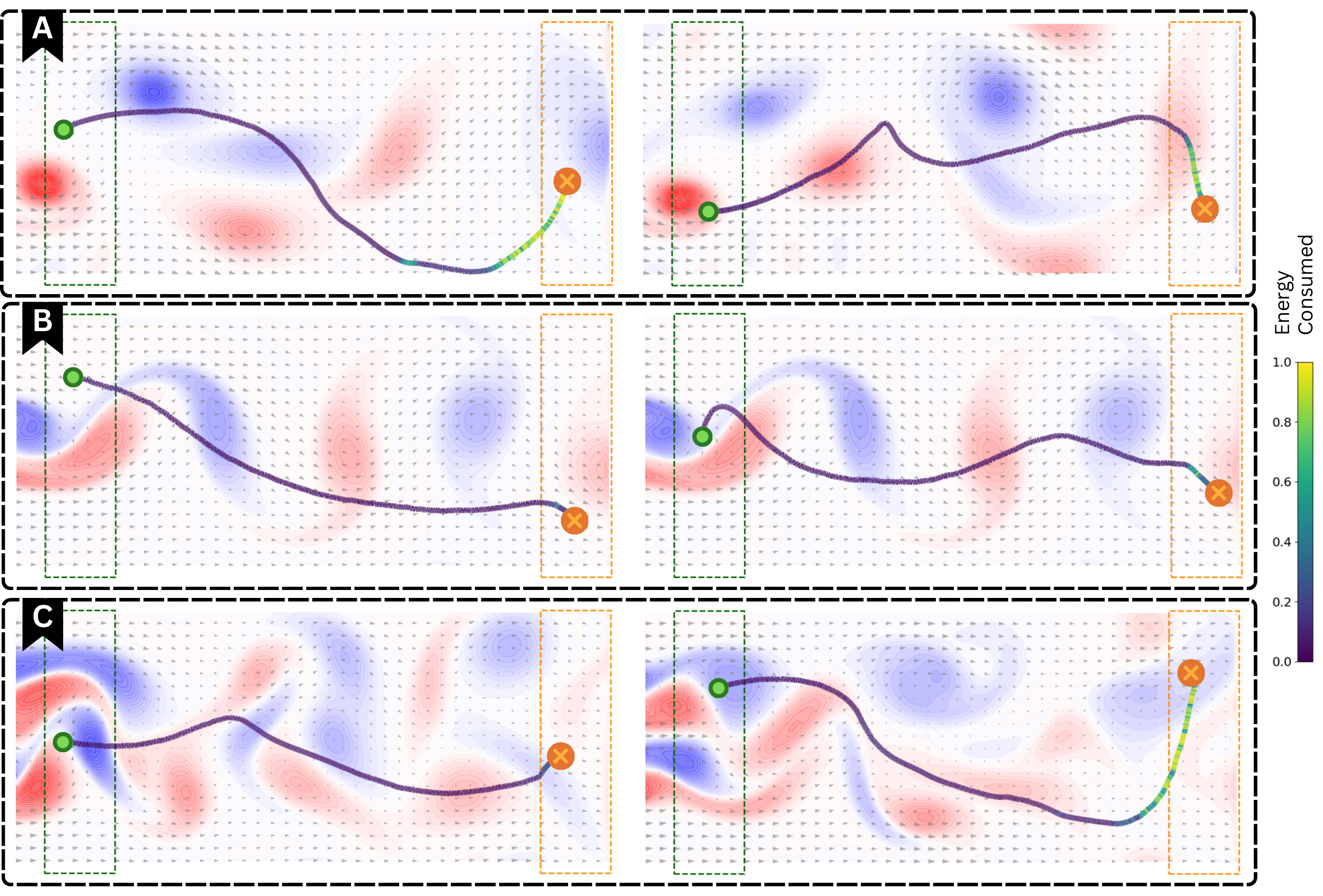}
    \caption{Sample agent trajectories in different flow environments. The color gradient along the path represents the energy used by the agent at each point. Note: the background flow shown in the figures corresponds to the initial flow at the start of each trial. \textbf{(A)} Trajectories in the oscillating single-cylinder environment. \textbf{(B)} Trajectories in the static single-cylinder environment (normal von Kármán vortex street). \textbf{(C)} Trajectories in the static double-cylinder environment.}
  \label{fig:test-env-trail}
\end{figure*}

 

\begin{figure*}[!ht]
  \centering
  \includegraphics[width=0.9\textwidth]{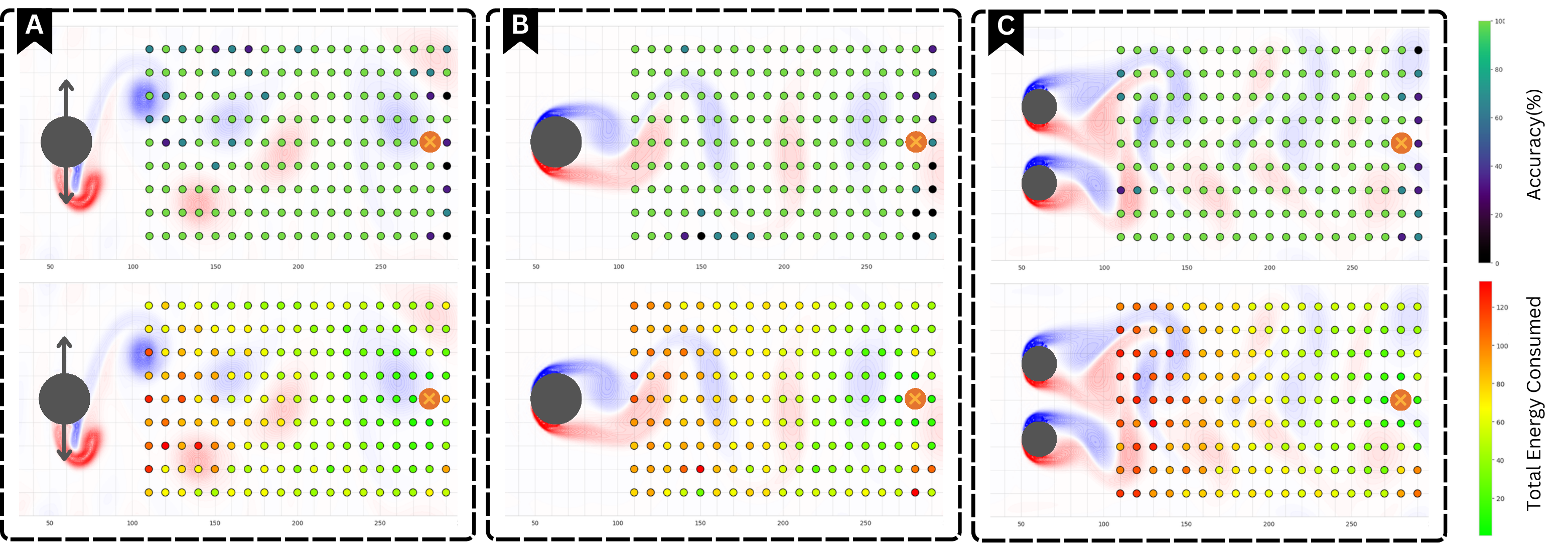}
  \caption{Navigation accuracy and energy efficiency across different agent spawn locations in various environments. The agent is spawned at positions on a $10 \times 10$ grid covering the entire environment, while the target is kept static at $(290, 50)$. Each spawn point value is averaged over five test runs with different flow initializations. For each environment, the top plot shows accuracy across spawn locations, and the bottom plot shows energy efficiency. Results are shown for \textbf{(A)} the oscillating single-cylinder environment, \textbf{(B)} the static single-cylinder environment, and \textbf{(C)} the static double-cylinder environment.}
  \label{fig:dotplot}
\end{figure*}

\section{RESULTS AND DISCUSSION}

\subsection{Navigation Results}

We evaluated navigation success in the same set of environments, with the addition of a new spawn layout - L-shaped spawn, as shown in Fig.~\ref{fig:setup}(C). This layout was not used during training, introducing new spawning scenarios that allow us to better measure the generalization of the policy. The results show high success rates across all environments and spawn layouts, as reported in Table~\ref{tab:nav-success}. Fig.~\ref{fig:test-env-trail} further illustrates sample trajectories where the agent effectively exploits local flow structures to reach the target. As observed from the energy gradient, the agent tries to catch and ride favorable currents towards the target to minimize energy consumption. In later stages, if it drifts off course, it maneuvers to recover and reach the target successfully. Fig.~\ref{fig:dotplot} further shows the navigation accuracy across the environments on a $10\times10$ grid covering the entire environment. Success rates are generally lower near the right boundary, where recovery is more difficult because movement in the target direction consistently opposes the flow directly. These findings confirm that the policy reliably navigates toward the target from a wide range of initial positions while depending only on single-point flow data.

\begin{table}[h!]
\centering
\begin{tabular}{cccc}
\hline
\textbf{Environment} & \makecell[c]{\textbf{Vertical} \\ \textbf{Spawn}} & \makecell[c]{\textbf{L-shaped} \\ \textbf{Spawn}} & \makecell[c]{\textbf{Grid Spawn} \\ \textbf{(10$\times$10)}} \\
\hline
\makecell[c]{\rule{0pt}{8pt} Oscillating Single \\ Cylinder} & \makecell[c]{95\%} & \makecell[c]{80\%} & \makecell[c]{94.33\%} \\[6pt]
\hline
\makecell[c]{Static Single \\ Cylinder} & \makecell[c]{100\%} & \makecell[c]{90\%} & \makecell[c]{92.98\%} \\[6pt]
\hline
\makecell[c]{Static Double \\ Cylinder} & \makecell[c]{70\%} & \makecell[c]{95\%} & \makecell[c]{95.3\%} \\
\hline
\end{tabular}
\caption{Navigation success rates (\%) across different environments and spawn layouts.}
\label{tab:nav-success}
\vspace{-1.5em}
\end{table}

\subsection{Energy-Efficiency Results}

To evaluate energy efficiency, we used grid based approach as a standard and widely accepted baseline. Many approaches discussed in the related work section employ A* or Dijkstra for initial path planning before optimizing the path to avoid obstacles using various techniques~\cite{trad-energy-eff-garau, zhangs-work, anis-work, trad-energy-eff-niu}. In our setup, where obstacles are not included, we compare against the first stage of their pipeline to provide a meaningful benchmark. This baseline benefits from having full knowledge of the flow field, making it a strong point of comparison. The heuristic for A* is the Euclidean distance from the agent’s position to the target. We created a dynamic graph spanning the entire flow field, where each node is connected to its eight neighbors. Edges are weighted based on the energy required to move from one node to another, accounting for the flow current. After each step, the flow values and edge weights are updated to reflect the current state. An RL baseline is also necessary to provide a comprehensive comparison. Therefore, we compared Khalasi with a RL implementation that followed the same experimental setup but without energy-based heuristics, namely the thrust penalty and surfing reward, in its reward function. While the setup differs, this approach aligns with the methodology presented by Jiao et al.~\cite{sensing-flow}.



\begin{table}[h!]
\centering
\begin{tabularx}{\linewidth}{cccc} 
\toprule
\textbf{Environment} 
 & \makecell{Oscillating\\Single Cylinder} 
 & \makecell{Static\\Single Cylinder} 
 & \makecell{Static\\Double Cylinder} \\
\midrule
\makecell{\textbf{Khalasi}\\\textbf{(Ours)}}
 & \makecell{$\mathbf{111.38 \pm 28.48}$}
 & \makecell{$\mathbf{80.97 \pm 18.93}$}
 & \makecell{$\mathbf{98.40 \pm 21.93}$} \\
\midrule
\makecell{\textbf{Graph Based}\\\cite{trad-energy-eff-garau, zhangs-work}\\\cite{anis-work, trad-energy-eff-niu}}
 & \makecell{$164.79 \pm 8.54$}
 & \makecell{$161.30 \pm 9.23$}
 & \makecell{$166.47 \pm 8.65$} \\
\midrule
\makecell{\textbf{RL Based}\\\cite{sensing-flow}}
 & \makecell{$183.68 \pm 25.42$}
 & \makecell{$174.26 \pm 17.31$}
 & \makecell{$179.72 \pm 24.23$} \\
\midrule
\makecell{\textbf{Mean}\\\textbf{Efficiency}}
 & \makecell{\textbf{35.89\%}}
 & \makecell{\textbf{51.67\%}}
 & \makecell{\textbf{43.07\%}} \\
\bottomrule
\end{tabularx}
\caption{Energy efficiency across different environments and approaches.}
\label{tab:energy-stats}
\end{table}






We achieved strong results, with the Khalasi pipeline delivering an overall energy efficiency improvement of 43.37\% compared to the baselines, despite relying solely on single-point data. On average, Khalasi outperforms A* by 41.68\% and naive RL by 46.72\%. Detailed results for both baselines are provided in Table~\ref{tab:energy-stats}. To better visualize how different spawn locations affect energy consumption, see Fig.~\ref{fig:dotplot}. A constant radial gradient centered around the goal is generally expected, as energy consumption typically increases with distance. In our results, we observe a similar pattern across environments, although the rate of change with distance is minimal; most spawn locations exhibit low to mid energy consumption. In the more complex double-cylinder environment, the gradient appears more linear, with slightly higher energy consumption on the left side compared to other environments.


\begin{figure*}[t]
  \centering
  \includegraphics[width=0.7\textwidth]{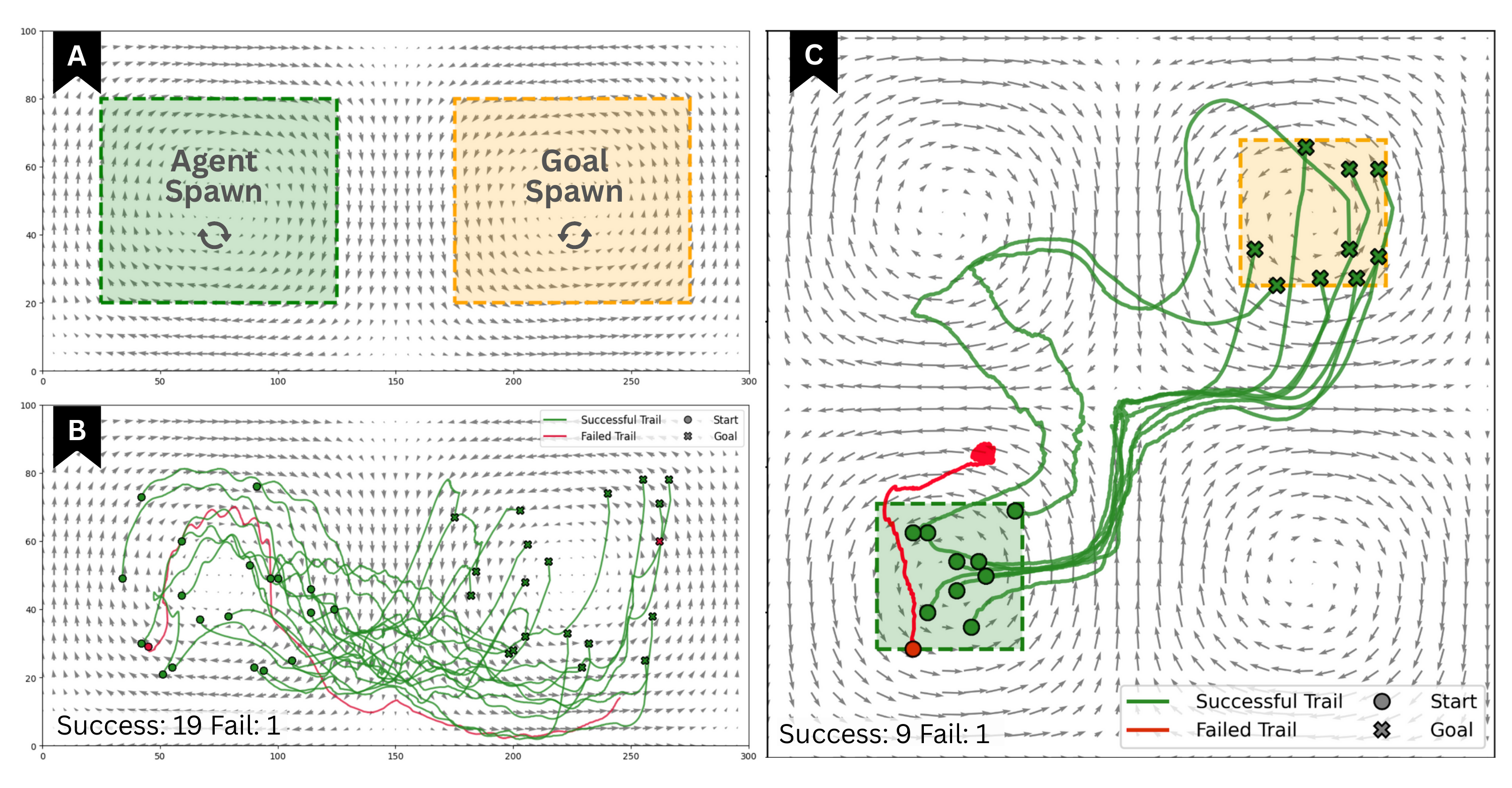}
  \caption{\textbf{(A)} Agent spawn region (green) and goal spawn region (orange) in the double-gyre environment. \textbf{(B)} Sample trajectories for double-gyre system testing. \textbf{(C)} Sample trajectories for quad-gyre system testing. Green lines indicate successful trails and red lines indicate failed ones in both cases.}
  \label{fig:test-generalize}
\end{figure*}


\subsection{Flow Generalization Results}

To evaluate the generalization ability of our model, we tested it on both real-world and synthetic flow data that were not included in the training.

\subsubsection{Double Gyre System}
Double-Gyre is a two-dimensional, time-periodic flow that serves as a simplified model of commonly found geophysical circulation patterns. Its structures have been extensively studied in the past~\cite{pastdoublegyre}. The stream function for the generalized gyre system is defined as
\[
\psi(x, y, t) = A \sin\big(N_x \pi f(x,t)\big) \sin(N_y \pi y)
\]
where
\[
f(x,t) = \epsilon \sin(\omega t)\, x^2 + x - 2\epsilon \sin(\omega t)\, x
\]
The background flow velocity field is then given by
\[
\mathbf{v}(x, y, t) =
\begin{bmatrix}
-\dfrac{\partial \psi}{\partial y} \\
\;\;\dfrac{\partial \psi}{\partial x}
\end{bmatrix}
=
\begin{bmatrix}
- A N_y \pi \sin\big(N_x \pi f(x,t)\big) \cos(N_y \pi y) \\
A N_x \pi \cos\big(N_x \pi f(x,t)\big) \sin(N_y \pi y) \dfrac{df}{dx}
\end{bmatrix}
\]

For double gyre we set $N_x = 2$ and $N_y = 1$ which are defined in the domain of $[0, 2] \times [0, 1]$. The parameter values are set to $\omega = 2\pi$ and $\varepsilon = 0.25$, while \textit{A} controls the flow velocity’s amplitude. We adjust \textit{A} to match the maximum flow velocity observed in the training environment. The environment has two regions for spawning the agent and the target, each located in a different gyre, as shown in Fig.~\ref{fig:test-generalize}A. No training was performed using the double-gyre flow. Despite this, the model generalized well during testing, achieving a navigation success rate of \textit{95\%}. Example subset of test trajectories are shown in Fig.~\ref{fig:test-generalize}B. The agent that spawns at far left of the spawn region, can be seen exploiting the current pattern to surf as long as it is beneficial. It then smoothly switches to the next gyre and continues surfing, tweaking its trajectory until it reaches the target.

\subsubsection{Quad Gyre System}
The same flow field defined above can be extended to create a quad-gyre system by using the domain $[0, 1] \times [0, 1]$ with $N_x = N_y = 2$. Similar to the double gyre, we set up two spawn areas for the agent and the target in different gyres arranged across from each other. No training was performed using the quad-gyre flow. Despite this, the model generalized well during testing and achieved a navigation success rate of 95\%. A subset of test trajectories is shown in Fig.~\ref{fig:test-generalize}C. Here, the model demonstrates an important ability that is, to recover from failed attempts. As seen in Fig.~\ref{fig:test-generalize}C, some agents initially drift in opposite directions along the $x$-axis while following the flow. Once the model identifies it has drifted too far, it corrects its path and moves toward the target.


\subsubsection{Real World Flow}
To validate our pipeline’s ability to navigate in real-world currents, we tested the model using data from the National Oceanic and Atmospheric Administration (NOAA). We selected a region in the North Atlantic with complex vortex structures that present significant navigation challenges. The data is defined over a $50 \times 50$ grid with a spatial resolution of approximately 6.6 km in both longitude and latitude. The area spans the bounding box between latitudes $39.360065^\circ$ and $41.130714^\circ$, and longitudes $-64.240000^\circ$ and $-61.920040^\circ$. The dataset includes hourly nowcasts over 21 days (from 2nd to 22nd November, 2021) totaling 504 data points.

\begin{figure}[h]
  \centering
  \includegraphics[width=0.8\columnwidth]{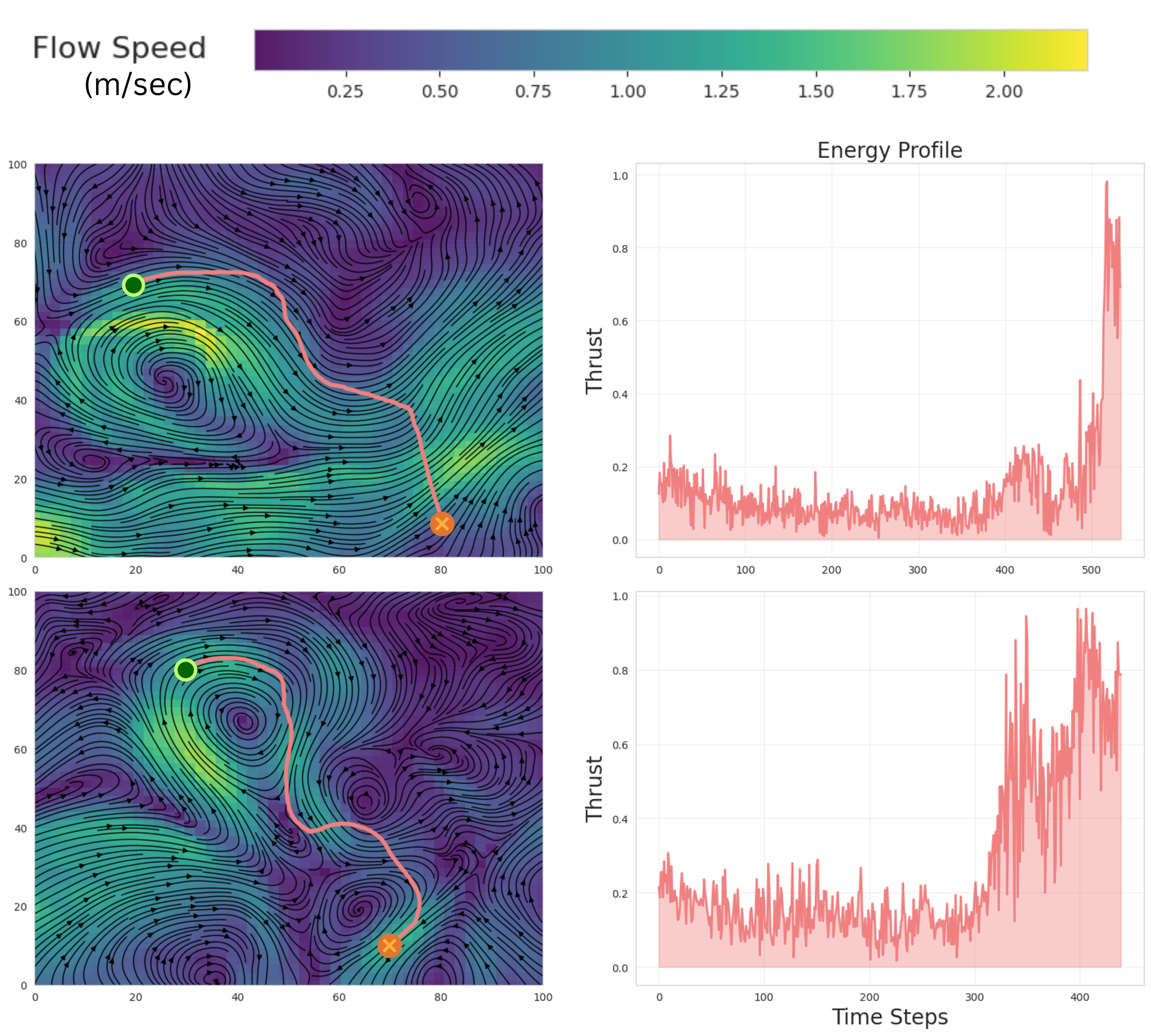}
    \caption{Example test trajectory and the corresponding energy consumption on the NOAA dataset. Note: the background flow shown in the figures corresponds to the initial flow at the start of each trial.}
  \label{fig:noaa}
\end{figure}

We interpolated the data both spatially and temporally by a factor of two, resulting in a grid size of $100 \times 100$ and enabling longer time-step experiments. The agent and goal were randomly spawned at different locations, with the criterion that they be at least 50 units apart in Euclidean distance. This test achieved a navigation success rate of 98\%. Sample trajectories with energy usage over time are shown in Fig.~\ref{fig:noaa}. In this case, the agent is underactuated relative to the flow velocity. As observed, the agent searches for flow patterns that help it progress toward the target and in later stages, applies higher thrust to maneuver effectively and reach the goal.

Compared to past works, our pipeline outperforms them by achieving better navigation accuracy using only its generalization capacity, without training on these specific flow.


\section{CONCLUSION AND FUTURE WORK}
We present Khalasi, a robust end-to-end pipeline for energy-efficient surface navigation in vortical flow fields, relying solely on local flow measurements. 
We extensively tested our policy and found that it consistently achieved high navigation success while increasing energy efficiency by $30-50\%$. The agent generalized to unseen synthetic flows and to real-world surface current data from NOAA without additional training. These results demonstrate that leveraging short-term history from on-board sensors combined with GPR based flow reconstruction and parallelized diverse flow training can significantly enhance the agent’s generalization performance. Overall, Khalasi offers a practical path toward longer ASV missions with reduced energy cost. Finally, we plan to release our code-base, along with an environment setup for others to develop, enhance, and use as benchmark for testing navigation systems in vortical flow fields.

We have an initial setup with an in-lab water tank and are working on creating consistent flow fields. We plan to deploy and test the pipeline on a physical setup to measure actual energy savings and system robustness in the field. Future work will also focus on real-world validation and narrowing the sim-to-real gap. We also plan to test small Vision-Language-Action (VLA) models in three-dimensional flows to measure their generalization capabilities across richer dynamics. 
Another improvement in the pipeline can be the integration of mission-level planning with battery models to optimize endurance and task allocation. Adaptive and sparse GPR and other error correction methods~\cite{gpr-error} can be investigated to reduce runtime while preserving uncertainty estimates. 




\bibliographystyle{IEEEtran}
\bibliography{IEEEabrv,mybibfile}

\end{document}